\documentclass[11pt]{article}

\usepackage[final]{acl}

\usepackage{times}
\usepackage{latexsym}
\usepackage{hyperref}
\usepackage[T1]{fontenc}
\usepackage[utf8]{inputenc}
\usepackage{microtype}
\usepackage{inconsolata}
\usepackage{graphicx}

\usepackage{booktabs}
\usepackage{ragged2e}
\usepackage{pgfplots}
\pgfplotsset{compat=1.17}
\usepackage{adjustbox}
\usepackage{tikz}
\usetikzlibrary{positioning, arrows.meta, shapes.geometric, fit, backgrounds}
\pgfdeclarelayer{background}
\pgfsetlayers{background,main}

\title{Structure-Guided Entity Resolution: Fine-Tuning LLMs for Robust Name Matching in Complex Linguistic Contexts\thanks{Accepted to ACL 2026. This is the author's preprint version; the final version will appear in the Proceedings of the 64th Annual Meeting of the Association for Computational Linguistics.}}

\author{
  Shivam Chourasia \\ Dream Sports \\ {\small\texttt{chshivam@utexas.edu}} \\
  \And
  Hitesh Kapoor \\ Dream Sports \\ {\small\texttt{hitesh.p.kapoor@gmail.com}} \\
  \And
  Nilesh Patil \\ Dream Sports \\ {\small\texttt{nilesh@urgrad.rochester.edu}} \\
}

\begin{document}
\maketitle

\begin{abstract}
Matching person names across heterogeneous records is a core challenge in entity resolution, especially within linguistically and culturally complex environments. Variations in naming conventions, inconsistent transliteration across scripts, and frequent data entry errors make it difficult to unify user identities, an essential requirement for Know Your Customer (KYC) compliance. While Large Language Models have shown promise in understanding natural language, they often struggle with the structured ambiguity present in such domain-specific settings. This paper introduces \textbf{Structure-Guided Entity Resolution (SGER)}, a novel framework that fine-tunes an LLM through a two-phase curriculum. The model is first trained to parse the grammatical and semantic structure of personal names, then optimized for the downstream task of binary entity matching. We evaluate SGER in the challenging context of Indian identity data, one of the most linguistically diverse and noisy environments globally. SGER achieves \textbf{99.02\% accuracy} and an \textbf{F1 of 0.994} on a held-out set of 50{,}000 real-world pairs, outperforming GPT-4o few-shot prompting and single-stage fine-tuning baselines. The system is fully deployed in production at Dream11, the world's largest fantasy sports platform, serving 250M+ users. Our results demonstrate that curriculum-guided training enables robust, high-precision entity resolution in real-world multilingual systems at scale.
\end{abstract}

\section{Introduction}

Entity Resolution (ER) is the task of identifying records that refer to the same real-world entity \citep{elmagarmid2007,fellegi1969}. In compliance-heavy domains such as KYC and AML, the quality of ER has direct operational and regulatory impact. This challenge is especially pronounced in culturally and linguistically diverse contexts, where names do not behave like stable keys. They vary with cultural convention, writing system, data entry practice, and the quirks of digitization.

The difficulties are especially visible in India. Naming conventions differ across regions and communities, and there is no single standard for which components count as first, middle, and last. A single individual's name may include patronymics, caste names, or village names as integral components, and these conventions vary by region and community. Honorifics and social suffixes such as \texttt{-bhai} and \texttt{-ji} appear in daily use and vanish in formal records. Scripts differ, and transliteration into Latin letters is not uniform across Devanagari, Bengali, Tamil, and other scripts \citep{iso15919_2001,steinberger2013jrcnames}. Operational pipelines add their own sources of variation: manual entry introduces typographical errors, and OCR blends tokens and drops whitespace. The same person may appear as ``Shubham Kumar Singh'' on a PAN card, ``Shubham K Singh'' in a bank record, and ``ShubhamKumarSingh'' in a utility bill scan. Table~\ref{table:challenges} lists concrete examples.

Our platform, Dream11, the world's largest fantasy sports platform serving over 250 million users, faces these variations daily. Missed links block legitimate users from completing verification, while false links can fail to flag fraudulent actors creating multiple accounts. Such fragmentation is a daily, high-volume challenge for platforms at scale. The limitations of traditional methods lead to high rates of both false positives and false negatives, carrying significant operational costs and regulatory risks. Classic techniques such as edit distance, Jaro-Winkler, and phonetic encodings provide useful signals but struggle with the full structure of the problem \citep{cohen2003,christophides2020}.

Large Language Models change the landscape because they capture regularities beyond local character edits. Prior work shows that fine-tuned LLMs can outperform conventional ER systems \citep{peeters2024,steiner2024,li2024leveraging,fan2024batcher,xin2024lora}. However, training directly on the binary decision forces the model to learn name structure and the decision boundary simultaneously, leaving performance untapped in domains where the structure is strong but implicit.

We propose \textbf{Structure-Guided Entity Resolution (SGER)}. We take Llama~3~8B and fine-tune it in two phases. Phase~1 teaches the model to map a raw name to a JSON schema with fields \texttt{first\_name}, \texttt{middle\_name}, and \texttt{last\_name}. Phase~2 starts from the Phase~1 checkpoint and trains a binary classifier. On a disjoint validation set of 50{,}000 pairs from Indian KYC workflows, SGER reaches 99.02\% accuracy and an F1 of 0.994. We describe our deployed system that serves this model in production, processing identity verification requests at scale with measurable business impact.

This paper's primary contribution is a highly accurate and deployed system for name-based entity matching in linguistically diverse environments that leverages a structured, curriculum-based fine-tuning strategy. We demonstrate that this approach allows the model to internalize the linguistic and cultural nuances of Indian names, achieving state-of-the-art accuracy and setting a new standard for entity resolution in complex, multilingual domains.

\renewcommand{\arraystretch}{1.3}
\begin{table*}[t!]
\centering
\caption{Examples of Name Matching Challenges in Multilingual Contexts (Indian Case Study)}
\label{table:challenges}
\resizebox{\textwidth}{!}{%
\begin{tabular}{@{}p{3cm}p{2.8cm}p{2.8cm}p{1.2cm}p{5.6cm}@{}}
\toprule
\textbf{Challenge Category} & \textbf{Example 1} & \textbf{Example 2} & \textbf{Match} & \textbf{Explanation} \\
\midrule
\raggedright Inverted or Merged Names & \raggedright Kirtan Singh & \raggedright SinghKirtan & Yes & Order of first and last names is swapped or concatenated. \\
\raggedright Spelling Inconsistencies & \raggedright Subham & \raggedright Shubham & Yes & Transliteration from regional scripts leads to inconsistent English spellings. \\
\addlinespace
\raggedright Phonetic Ambiguity & \raggedright Vipin & \raggedright Bipin & No & Similar sounding names that refer to different individuals. \\
\raggedright Truncations \& Abbreviations & \raggedright Rajeshk & \raggedright Rajesh Kumar & Yes & Informal records often use shortened or abbreviated names. \\
\addlinespace
\raggedright Patronymic Variants & \raggedright Rajesh So Hari & \raggedright Hari & No & Father's name appears as a relational suffix (son of). \\
\raggedright Honorifics \& Postfixes & \raggedright Rameshbhai Patel & \raggedright Ramesh Patel & Yes & Suffixes like \texttt{-bhai}, \texttt{-ji}, or honorifics are inconsistently applied. \\
\bottomrule
\end{tabular}%
}
\end{table*}

\section{Related Work}

The literature on name matching spans basic string comparison, phonetic heuristics, classical supervised learning, and modern neural approaches. We summarize relevant strands and place SGER within that landscape.

\paragraph{Heuristics and rules:}
Edit distance, Jaro-Winkler, and phonetic encodings capture surface similarity but struggle with token reordering, merged whitespace, and cross-script transliteration \citep{cohen2003,christen2012book}. Phonetic algorithms such as Soundex and Metaphone are tuned for English pronunciation and are ill-suited for Indian languages, where morphological and cross-script variation dominates \citep{mhaske2022}.

\paragraph{Classical ML:} Before deep learning, supervised models using string similarity features and handcrafted rules outperformed individual heuristics but required extensive feature engineering \citep{christophides2020}.

\paragraph{Deep learning and PLMs:} Pre-trained language models like BERT \citep{devlin2019bert}, when fine-tuned for ER, significantly improve over previous methods by capturing richer semantic information \citep{li2023}.

\paragraph{LLMs for entity matching:} The survey by \citet{peeters2023} maps the landscape, while \citet{steiner2024} demonstrate that fine-tuning is highly effective. \citet{huang2024} proposes a relation-based approach for high-stakes tasks. Our SGER methodology implicitly addresses this concern: by first training the model to understand structural relationships between name components (Phase~1), we equip it to disambiguate difficult cases in the downstream binary setting. We are the first to empirically validate a two-phase curriculum learning framework \citep{feng2023,soviany2021curriculum} tailored to name-based entity resolution in complex real-world settings.

We argue that in domains with implicit ``grammar'' (e.g., culturally diverse names), standard fine-tuning is suboptimal, as it forces the model to simultaneously learn structural patterns and perform classification. Our curriculum learning strategy separates these concerns, leading to improved performance. While prior work has explored aspects of name processing in Indic languages \citep{bahad2024}, none have adopted a structured, multi-phase training paradigm that incrementally builds linguistic understanding before downstream decision-making.

\section{Methodology: Curriculum-Based Entity Resolution}

To tackle the challenges outlined above, we developed the SGER system. The core of our methodology is a two-phase, curriculum-based fine-tuning approach that equips a pre-trained LLM with specialized knowledge of name structures before training it on the matching task. Our central hypothesis is that by decoupling structural understanding from the matching task, the model develops a more robust and generalizable internal representation of names, providing a powerful inductive bias for the downstream classification task.

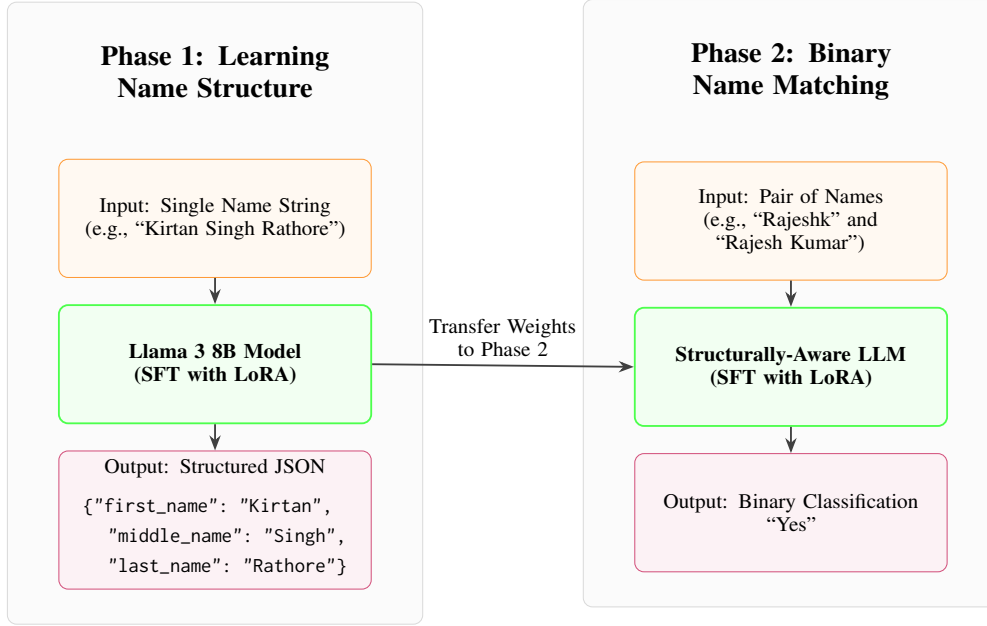
\begin{figure*}[t!]
\centering
\begin{adjustbox}{width=0.82\textwidth,center}
\begin{tikzpicture}[
node distance=1.8cm and 3.5cm,
llm/.style={rectangle, draw=green!70, fill=green!5, text width=4.5cm, text centered, rounded corners, minimum height=1.8cm, font=\footnotesize\bfseries, thick},
input_box/.style={rectangle, draw=orange!80, fill=orange!5, text width=4.5cm, text centered, rounded corners, minimum height=1.8cm, font=\small},
output_box/.style={rectangle, draw=purple!70, fill=purple!5, text width=4.5cm, text centered, rounded corners, minimum height=1.8cm, font=\small},
arrow/.style={-Stealth, thick, draw=black!75},
panel_label/.style={font=\bfseries\large, text width=5cm, align=center},
background_box/.style={draw=black!15, fill=gray!3, fit=#1, inner sep=15pt, rounded corners}
]

\node (phase1_title) [panel_label] {Phase 1: Learning\\Name Structure};
\node (input1) [input_box, below=0.8cm of phase1_title] {Input: Single Name String \\ (e.g., ``Kirtan Singh Rathore'')};
\node (llm1) [llm, below=0.4cm of input1] {Llama 3 8B Model \\ (SFT with LoRA)};
\node (output1) [output_box, below=0.4cm of llm1, align=center] {
    Output: Structured JSON \\[0.5em]
    \begin{tabular}{@{}l@{}}
    \ttfamily\footnotesize\{"first\_name": "Kirtan", \\
    \ttfamily\footnotesize\phantom{\{}\ \ "middle\_name": "Singh", \\
    \ttfamily\footnotesize\phantom{\{}\ \ "last\_name": "Rathore"\}
    \end{tabular}
};

\node (phase2_title) [panel_label, right=of phase1_title] {Phase 2: Binary\\Name Matching};
\node (input2) [input_box, below=0.8cm of phase2_title] {Input: Pair of Names \\ (e.g., ``Rajeshk'' and \\ ``Rajesh Kumar'')};
\node (llm2) [llm, below=0.4cm of input2] {Structurally-Aware LLM \\ (SFT with LoRA)};
\node (output2) [output_box, below=0.4cm of llm2] {Output: Binary Classification \\ ``Yes''};

\begin{pgfonlayer}{background}
    \node [background_box={(phase1_title)(input1)(llm1)(output1)}] {};
    \node [background_box={(phase2_title)(input2)(llm2)(output2)}] {};
\end{pgfonlayer}

\draw [arrow] (input1) -- (llm1);
\draw [arrow] (llm1) -- (output1);
\draw [arrow] (input2) -- (llm2);
\draw [arrow] (llm2) -- (output2);
\draw [arrow] (llm1.east) -- node[above, font=\small, text width=3cm, align=center] {Transfer Weights \\ to Phase 2} (llm2.west);

\end{tikzpicture}
\end{adjustbox}
\caption{Structure-Guided Entity Resolution (SGER) methodology. Phase~1 fine-tunes Llama~3~8B to parse names into structured JSON using SFT with LoRA. The learned weights transfer to Phase~2, where the model performs binary name matching.}
\label{fig:SGER_methodology}
\end{figure*}

An overview of the two-phase training pipeline is illustrated in Figure~\ref{fig:SGER_methodology}. Phase~1 focuses on name structure understanding, where the model is fine-tuned to output structured JSON representations from full name strings. In Phase~2, the fine-tuned model is used as a starting point for binary name matching, where it classifies whether two name variants refer to the same individual.

\subsection{System Architecture}

SGER uses Meta's Llama~3~8B as the base \citep{llama32024}. The choice balances accuracy with cost: the model is large enough to capture regularities in Indian names and small enough to fine-tune and serve efficiently. Llama~3~8B is open-source, well-documented, and widely adopted in both academic and industrial contexts, making it a natural and reproducible choice. Our task is linguistically complex but narrow in scope and highly structured. This allows us to use a model of modest size without compromising on accuracy. As our results show, the curriculum-based fine-tuning pipeline enables Llama~3~8B to achieve near-perfect accuracy, demonstrating that larger models are not necessarily required for specialized entity resolution tasks. The input at inference is a pair of name strings, and the output is a single token: \texttt{``Yes''} or \texttt{``No''}. The novelty lies in the training path rather than in a new network component.

\subsection{Phase 1: Name Structure Understanding}
\label{sec:phase1}
Phase~1 teaches the model the internal grammar of names through a supervised mapping from a single name string to a JSON object listing its main components.

\textbf{Input:} A single name string, e.g., ``Kirtan Singh Rathore''. Inputs are space-trimmed and lowercased so that the model encounters the same quirks that appear in real data.

\textbf{Output:} A structured JSON object:
\begin{verbatim}
{
  "first_name": "Kirtan",
  "middle_name": "Singh",
  "last_name": "Rathore"
}
\end{verbatim}

\textbf{Training Data:} We constructed a dataset of approximately 10{,}000 Indian names, manually annotated with their structural components. The names were sampled from anonymized identity records and augmented with synthetically generated names to improve coverage and diversity. Particular attention was given to capturing a broad spectrum of regional, linguistic, and cultural naming patterns, including non-standard spellings, varied name orders, and merged or abbreviated forms. This dataset is distinct from the binary name-matching pairs described in Section~\ref{sec:data} and serves as an upstream pretraining corpus that equips the model with structural awareness prior to fine-tuning on the matching task.

\textbf{Optimization:} We apply supervised fine-tuning with LoRA on Llama~3~8B. The task encourages the model to build a stable internal representation of how names are composed, lowering the difficulty of Phase~2. We use mixed precision training and early stopping.

By completing this task, the model learns to identify common first names, family names, and middle names (like ``Kumar'' or ``Lal''). It also learns to handle structural variations, such as recognizing that ``SinghKirtan'' is likely a merged form of a first and last name. This phase acts as a form of domain-specific pre-training, equipping the model with an internal representation of name semantics before the binary matching task.

\subsection{Phase 2: Binary Name Matching}
\label{sec:phase2}
Phase~2 continues from the Phase~1 checkpoint and trains a classifier over pairs of names.

\textbf{Input:} Each instance contains a brief instruction, a few-shot block with examples, and the target pair:
\begin{quote}
\texttt{\textit{[Instruction]}}\\
\texttt{\textit{[Few-Shot Examples]}}\\
\texttt{Name 1: "}\textit{A}\texttt{" | Name 2: "}\textit{B}\texttt{" -> Match?}
\end{quote}
\textbf{Prompt Design:} The instruction and few-shot prompts were crafted with the risk operations team to reflect the full range of real-world name ambiguities seen in production. This includes handling of abbreviations, merged tokens, transliteration inconsistencies from regional scripts, reordered components, and honorific or suffix variations. The same instruction template was used across all few-shot and fine-tuned model setups, ensuring consistency during training and inference.

\textbf{Output:} A single classification token: either \texttt{``Yes''} or \texttt{``No''}, indicating whether the two input names refer to the same individual. At inference, we extract the model's scores for the answer tokens \texttt{``Yes''} and \texttt{``No''}, apply a softmax to obtain the probability of a match, and classify a pair as positive when this probability exceeds a decision threshold selected to maximize F1 on the validation set. This yields deterministic, calibrated binary decisions.

\textbf{Training Data:} SFT with LoRA on labeled name pairs (Section~\ref{sec:data}). The prompt template is used during training to match the evaluation setup.

\section{Experimental Setup}

\subsection{Data}
\label{sec:data}
The evaluation corpus is proprietary and drawn from our platform's historical KYC verification workflows. Records contain name pairs only; no other PII fields are present.

We begin with 20{,}000 labeled name pairs curated from our past operations and expand to more than 50{,}000 examples with three domain-informed augmentation strategies.

Our augmentation philosophy draws from computer vision, where
transforms such as horizontal flips, geometric rotations, and noise
injection encode known invariances of the visual domain
\citep{shorten2019}. We apply the same principle to names: each
augmentation strategy encodes a specific invariance relevant to the
name-matching domain.
\begin{itemize}
    \item \textbf{Pair Swapping}, analogous to mirroring in CV, adds for each pair (name1, name2) the swapped pair (name2, name1) to ensure the model is order-invariant.
    \item \textbf{Component Permutation}, analogous to geometric transforms, permutes the constituent parts (first, middle, last name) of names to create structurally valid but novel examples.
    \item \textbf{Random Space Removal}, analogous to noise injection, randomly removes spaces between name parts (e.g., ``Kirtan Singh'' becomes ``KirtanSingh'') to simulate common data entry errors.
\end{itemize}

Evaluation uses a disjoint held-out set of 50{,}000 real-world pairs drawn from our historical production data and manually labeled by our in-house risk operations team. Disjointness is enforced at the individual name level: no name appearing in any test pair occurs in any training pair, whether original or augmented, preventing data leakage across splits.
This held-out set is deliberately 2.5$\times$ the base training corpus to provide high confidence in production performance, as the system operates in a compliance-critical identity verification pipeline.
The impact of each augmentation strategy is quantified in our
ablation study (Section~\ref{sec:ablation}).
\subsection{Baselines}
We compare SGER with five systems:
\begin{itemize}
    \item \textbf{Levenshtein Similarity:} Fuzzy string matching with threshold 0.8 tuned for F1 \citep{cohen2003,christen2012book}.
    \item \textbf{BERT (Fine-Tuned):} Multilingual \texttt{bert-base-multilingual-cased} with a classification head over [CLS] on concatenated names.
    \item \textbf{GPT-4o (Few-Shot):} Prompted with the same instruction template and curated examples as SGER, without fine-tuning.
    \item \textbf{LLM (Few-Shot, Llama 3 8B):} The same few-shot setup on the base Llama~3~8B model.
    \item \textbf{LLM (SFT, Llama 3 8B):} A one-stage fine-tune directly
on the binary task. We report variants with and without data
augmentation to isolate its contribution.
\end{itemize}

\subsection{Training Details}
All experiments use Llama~3~8B Instruct as the base model. Both phases rely on SFT with LoRA \citep{hu2021lora}: rank~4, scaling factor \texttt{alpha}~8, mixed precision (\texttt{fp16}). We adopt default hyperparameters from PEFT and Hugging Face Transformers with early stopping on validation F1. Training is conducted on a single NVIDIA A100 GPU with 80GB memory. Inputs are space-trimmed and lowercased.

\subsection{Evaluation Metrics}
We report accuracy, precision, recall, and F1. High precision reduces false links (compliance risk); high recall reduces friction for legitimate users. We use F1 for early stopping.

\section{Results and Discussion}

\subsection{Quantitative Analysis}
\begin{table}[h]
\centering
\caption{Performance Comparison of Name Matching Models}
\label{table:results}
\resizebox{\columnwidth}{!}{%
\begin{tabular}{lcccc}
\hline
\textbf{Model / Method} & \textbf{Accuracy} & \textbf{Precision} & \textbf{Recall} & \textbf{F1-Score} \\
\hline
Levenshtein (Th=0.8)              & 57.4\% & 75.1\% & 70.2\% & 0.726 \\
BERT (Fine-Tuned)                 & 69.1\% & 81.2\% & 80.1\% & 0.806 \\
GPT-4o (Few-Shot)                 & 85.4\% & 90.1\% & 92.1\% & 0.911 \\
LLM (Few-Shot, Llama 3 8B)        & 74.3\% & 85.2\% & 82.6\% & 0.839 \\
LLM (SFT, Llama 3 8B)             & 91.2\% & 93.3\% & 95.9\% & 0.946 \\
LLM (SFT + Aug., Llama 3 8B)      & 95.7\% & 97.6\% & 97.0\% & 0.973 \\
\textbf{SGER (Our Method)}        & \textbf{99.02\%} & \textbf{99.95\%} & \textbf{98.9\%} & \textbf{0.994} \\
\hline
\end{tabular}%
}
\end{table}

SGER outperforms all baselines by a significant margin. Classical methods like Levenshtein struggle with structural and phonetic variations. GPT-4o performs strongly in the few-shot setting but still falls short, highlighting the value of domain-specific fine-tuning. The few-shot Llama~3 model underperforms its fine-tuned variants, confirming that in-domain supervision is essential. SGER's curriculum combined with data augmentation pushes F1 from 0.946 to 0.994 over the single-stage baseline, demonstrating that curriculum learning enables production-grade entity matching.

\subsection{Ablation Study}
\label{sec:ablation}
The rows in Table~\ref{table:results} are ordered to isolate the
contribution of each pipeline component, starting from a vanilla
single-stage SFT baseline and adding one component at a time.

\paragraph{Effect of Data Augmentation.}
Training on the original 20K curated pairs alone leaves the model
under-exposed to structural variants that appear frequently in
production data. Expanding to 50K+ examples with the augmentation
strategies described in Section~\ref{sec:data} raises the F1 from
0.946 to 0.973. Without these augmentations, the model memorizes
surface patterns of the curated set but fails to generalize to the
long tail of distortions encountered in production.

\paragraph{Effect of Curriculum Learning.}
Adding the Phase~1 structural pre-training step
(Section~\ref{sec:phase1}) before Phase~2 matching yields a further
gain. The curriculum helps most on pairs combining multiple
distortions, for example an abbreviated and reordered name, where
Phase~1 stabilizes the model's internal representation of name
components before the matching decision. Together, data augmentation
and curriculum learning are complementary and their combined effect
brings SGER to an F1 of 0.994.

\subsection{Error Analysis}
Manual review of misclassified pairs revealed two primary failure modes:
\begin{itemize}
    \item \textbf{Ambiguous Phonetic Cases:} Names phonetically similar but spelled differently (e.g., \texttt{"Saurabh"} vs.\ \texttt{"Sorab"}) were sometimes classified as non-matches due to inconsistent transliteration or regional pronunciation norms.
    \item \textbf{Complex Compound Errors:} Cases combining multiple simultaneous distortions (missing whitespace, inverted order, and OCR corruption) remain challenging. For instance, \texttt{``SubhamKshr''} vs.\ \texttt{``KishoreShubham''} was misclassified due to concurrent whitespace, ordering, and character-level errors.
\end{itemize}

\section{Production Deployment and Impact}

SGER is fully deployed in our production identity verification pipeline, serving over 250 million users on Dream11.

\subsection{Deployment Architecture}
We deploy the model on a Kubernetes-based architecture providing horizontal scalability, high availability, and fault tolerance. Inference runs on 3$\times$ NVIDIA L4 GPUs. We serve the model using the vLLM framework \citep{vllm2023}, which enables efficient token-level scheduling and high-throughput batching.

\subsection{System Performance}
Our deployed system handles up to \textbf{10{,}000 requests per minute} (RPM) with a \textbf{P99 latency of 120 milliseconds}. This level of throughput is critical during major sporting events, when our user onboarding traffic spikes significantly.

\subsection{Business Impact}
Prior to SGER, heuristic-based matching routed ambiguous cases to a manual review queue.

\textbf{Elimination of manual review.} With 99.95\% precision, SGER enables fully automated match/no-match decisions. The rare false negatives ($<$1\% of users) are resolved through customer service.

\textbf{Cost savings.} Prior to SGER, ambiguous cases were routed to a third-party vendor for manual verification. Eliminating this dependency translates to over \textbf{\$500K in annual cost savings} from reduced operational overhead.

\section{Conclusion}

We presented SGER, a curriculum-based approach to name-only entity
resolution that first trains an LLM to understand name structure
before fine-tuning it on binary matching. On Indian KYC data, SGER
achieves a state-of-the-art F1 of 0.994, eliminating manual review
and delivering over \$500K in annual savings. Its successful
deployment in our platform's production pipeline underscores its
robustness, scalability, and real-world value.

While our experiments focus on Indian KYC data, SGER is broadly
applicable to other settings with rich linguistic diversity,
irregular naming, or noisy inputs. Applications such as multilingual
identity systems, global e-commerce, and cross-border fraud
detection can benefit from this approach. Our error analysis suggests
several promising directions: combining SGER with phonetic algorithms
for ambiguous transliteration cases and extending to a multimodal system
that processes identity document images to resolve OCR-related noise
at their origin \citep{sun2025robustner}.

\section*{Ethical Considerations}

The dataset used in this work is proprietary and drawn from our platform's anonymized KYC records. No personally identifiable information was used in model evaluation or is reported in this paper. All data processing was conducted under our internal data protection agreements. The deployed system is designed to reduce bias in identity verification by handling diverse naming conventions more accurately than rule-based approaches.

\bibliography{custom}

\end{document}